\documentclass[10pt,twocolumn,letterpaper]{article}

\usepackage{cvpr}
\usepackage{times}
\usepackage{epsfig}
\usepackage{graphicx}
\usepackage{amsmath}
\usepackage{amssymb}

\usepackage[pagebackref=true,breaklinks=true,letterpaper=true,colorlinks,bookmarks=false]{hyperref}

\cvprfinalcopy 

\def\cvprPaperID{****} 

\ifcvprfinal\fi
\begin{document}

\title{Dynamic Multimodal Sentiment Analysis: Leveraging Cross-Modal Attention for Enabled Classification}

\author{Hui, Lee\\
{\tt\small hlee3032@gatech.edu}
\and
Singh, Suniljit\\
{\tt\small ssingh932@gatech.edu}
\and
Yong Siang, Ong\\
{\tt\small yong32@gatech.edu}
}

\maketitle

\begin{abstract}
    This paper explores the development of a multimodal sentiment analysis model that integrates text, audio, and visual data to enhance sentiment classification. The goal is to improve emotion detection by capturing the complex interactions between these modalities, thereby enabling more accurate and nuanced sentiment interpretation. The study evaluates three feature fusion strategies—late stage fusion, early stage fusion, and multi-headed attention—within a transformer-based architecture. Experiments were conducted using the CMU-MOSEI dataset, which includes synchronized text, audio, and visual inputs labeled with sentiment scores. Results show that early stage fusion significantly outperforms late stage fusion, achieving an accuracy of 71.87\%, while the multi-headed attention approach offers marginal improvement, reaching 72.39\%. The findings suggest that integrating modalities early in the process enhances sentiment classification, while attention mechanisms may have limited impact within the current framework. Future work will focus on refining feature fusion techniques, incorporating temporal data, and exploring dynamic feature weighting to further improve model performance.
\end{abstract}

\section{Introduction}

This project aimed to develop a multimodal sentiment analysis model~\cite{soleymani2017surveymsa} that integrates text, audio and visual data to improve sentiment classification. The goal was to improve how emotions are classified, enabling computers to interpret human communication more effectively. By leveraging all three modalities, the model seeks to capture a richer and more nuanced understanding of sentiment compared to existing methods.\\

Current approaches of feature fusion typically include concantenation and early or late fusion strategies~\cite{lai2023msasurvey}, which may overlook the complex interactions between modalities. Advanced methods like attention mechanisms improve integration, they often fail to fully capture the intricate relationships between the modalities, particularly over time. These limitations leave room for improvement in understanding how words, tone, and expressions interact to convey sentiment.\\

If successful, this project could significantly improve applications such as social media monitoring, customer feedback analysis, and human-computer interaction. For example, it could help businesses more accurately gauge customer satisfaction or make virtual assistants more responsive and empathetic.\\

The project utilized the processed CMU-MOSEI (Multimodal Opinion Sentiment and Emotion Intensity) dataset~\cite{liang2023multizoo}, which provides multimodal data with text, audio, and video elements. This dataset is labeled with the correct sentiments, making it ideal for modeling how different modalities interact. Its comprehensive nature enables the team to explore dynamic and context-aware sentiment classification effectively.
\section{Approach}

The project aims to improve multimodal sentiment classification by leveraging cross-modal attention mechanisms. The core idea is to develop a model that dynamically selects relevant features from each modality based on context and temporal dependencies, offering more accurate sentiment predictions.

\subsection{Related Work}
Several existing models inform our approach to multimodal sentiment analysis, yet they each have limitations that our project aims to address. \textbf{MAG-BERT}~\cite{rahman2020magbert} utilizes a vector embedding structure to sequence multimodal inputs for BERT, employing a multimodal adaptive gate fusion strategy. However, it may not fully capture the nuanced interactions among modalities over time. \textbf{TIMF}~\cite{sun2021timf} implements feature fusion through a tensor fusion network, preserving relationships among modalities for nuanced sentiment understanding. Yet, it lacks dynamic adaptability in feature selection. \textbf{DISRFN}~\cite{he2022DISFRN} focuses on obtaining domain-separated representations, dynamically fusing them through a hierachical graph network. While effective for handling data variability, it does not specifically address the integration of temporal dynamics inherent in multimodal data. \textbf{TEDT}~\cite{wang2022tedt} adopts a transformer-based encoding-decoding approach using a modality reinforcement cross-attention module, yet it may not fully leverage the inter-dependencies among modalities. Our project aims to address these by focusing on dynamic adaptability in feature selection across modalities and improved temporal integration.

\subsection{Data Collection}
The primary dataset used in this research is the CMU-MOSEI dataset, which provides multimodal data for sentiment analysis. The dataset includes synchronized text, audio, and visual inputs, each labeled with sentiment scores on a scale from -3 to +3. These continuous sentiment scores are then discretized into three categories to simplify the classification task:
\begin{itemize}
  \item Negative Sentiment: Sentiments ranging from [-3, -1)
  \item Neutral Sentiment: Sentiments ranging from [-1, 1]
  \item Positive Sentiment: Sentiments ranging from (1, 3]
\end{itemize}

\subsection{Preprocessing in CMU MOSEI dataset}
The CMU-MOSEI dataset has already been preprocessed to align the text, audio, and visual modalities. Key features have also been extracted:
\begin{itemize}
    \item \textbf{Text:} Transcriptions are provided, and word embeddings are extracted using GloVe (840B tokens). The text and audio are aligned at the phoneme level using the P2FA forced alignment model.
    \item \textbf{Audio: } Features such as MFCCs, pitch, and glottal sound parameters are extracted using COVAREP.
    \item \textbf{Visual: } Frames are extracted at 30Hz, with faces detected using MTCNN. FACS is used to extract facial action units, and Emotient FACET is used for emotion recognition. Additional facial features, including landmarks and head pose, are captured using OpenFace, and face embeddings are obtained from DeepFace and FaceNet.
\end{itemize}

\subsection{Modality Transformer Architecture}
\label{subsec:modality_transformer_architecture}
Each modality is processed using a shared ModalityTransformer architecture, consisting of the following components: 1) a linear layer that transforms features into a common model dimension, 2) a Transformer Encoder with positional encoding, and 3) a classification layer for sentiment analysis.\\

\subsection{Feature Fusion Approach}
Three different feature fusion strategies are explored in our research.

\subsubsection{Approach 0: Late Stage Feature Fusion}
\label{subsec:a0}
In this approach, the classifiers trained for each modality (text, audio, and visual), as described in Section~\ref{subsec:modality_transformer_architecture}, are utilized to make final sentiment predictions through a majority voting mechanism. This method functions similarly to an ensemble learning technique, combining the strengths of individual classifiers. By doing so, it establishes a baseline for performance comparison with more advanced and integrated multimodal approaches.

\subsubsection{Approach 1: Early Stage Feature Fusion}
\label{subsec:a1}
Features extracted from each modality are concatenated at the final hidden state and then passed through a classifier to predict the sentiment. This approach directly combines the features before classification, enabling the model to learn a shared representation of the modalities.

\subsubsection{Approach 2: Multi-headed Attention}
\label{subsec:a2}
This approach builds upon Approach 1 by adding an attention layer that learns to weigh the importance of each modality’s features dynamically. This allows the model to focus on the most relevant features at each step in the sentiment analysis process, improving the integration of multimodal data.

\section{Experiments and Results}


\subsection{Experimental Setup}
The three approaches and the models for each modality were trained, validated and tested on the CMU-MOSEI dataset. The models were trained using a cross-entropy loss function and optimized using the Adam optimizer. All experiments were implemented using PyTorch. To enable comparison, each approach was trained and evaluated using the accuracy metric.\\

\subsection{Results}
The accuracy results on the test dataset for the three models described in ~\ref{subsec:modality_transformer_architecture} are compiled in Table \ref{tab:each_modality_results}.

\begin{table}[h!]
\centering
\renewcommand{\arraystretch}{1.3} 
\begin{tabular}{|l|c|}
\hline
\textbf{Modality} & \textbf{Accuracy Metric} \\ \hline
Video & 65.69\% \\ \hline
Audio & 65.84\% \\ \hline
Text & 69.36\% \\ \hline
\end{tabular}
\caption{Accuracy metrics for each of the three modality models.}
\label{tab:each_modality_results}
\end{table}

Similarly, the results of the three approaches described in ~\ref{subsec:a0},~\ref{subsec:a1} and ~\ref{subsec:a2} are compiled in Table \ref{tab:results}.

\begin{table}[h!]
\centering
\renewcommand{\arraystretch}{1.3} 
\begin{tabular}{|l|c|}
\hline
\textbf{Approach Type} & \textbf{Accuracy Metric} \\ \hline
A0: Late Stage Feature Fusion & 66.23\% \\ \hline
A1: Early Stage Feature Fusion & 71.87\% \\ \hline
A2: Multi-headed Attention & 72.39\% \\ \hline
\end{tabular}
\caption{Accuracy metrics for the different approach types.}
\label{tab:results}
\end{table}

\subsubsection{Model for Each Modality}

\begin{figure}[htp]
    \centering
    \includegraphics[width=8cm]{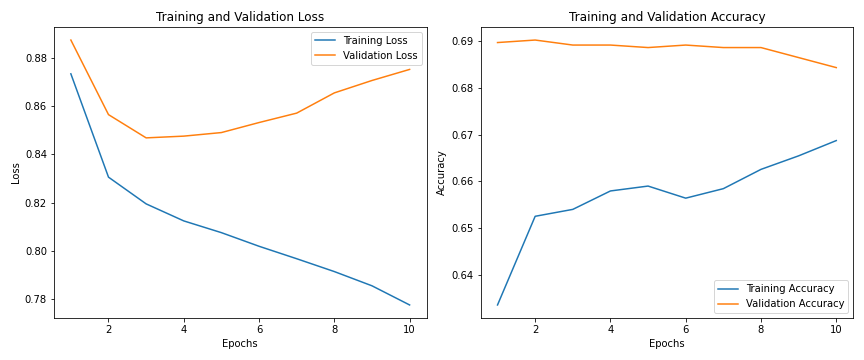}
    \caption{Training and validation loss and accuracy for video model}
    \label{fig:video_graph}
\end{figure}

\begin{figure}[htp]
    \centering
    \includegraphics[width=8cm]{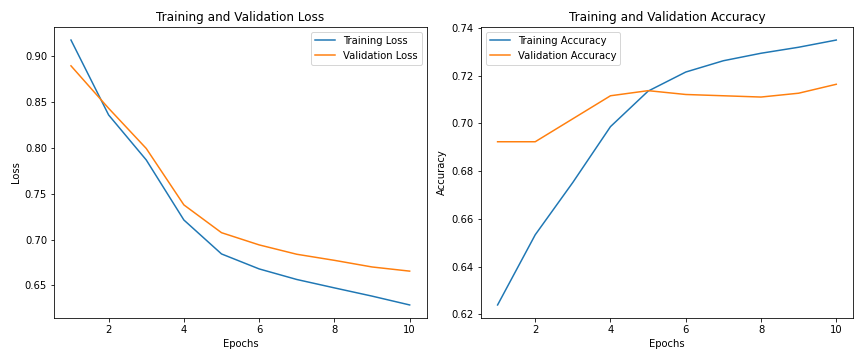}
    \caption{Training and validation loss and accuracy for text model}
    \label{fig:text_graph}
\end{figure}

\begin{figure}[htp]
    \centering
    \includegraphics[width=8cm]{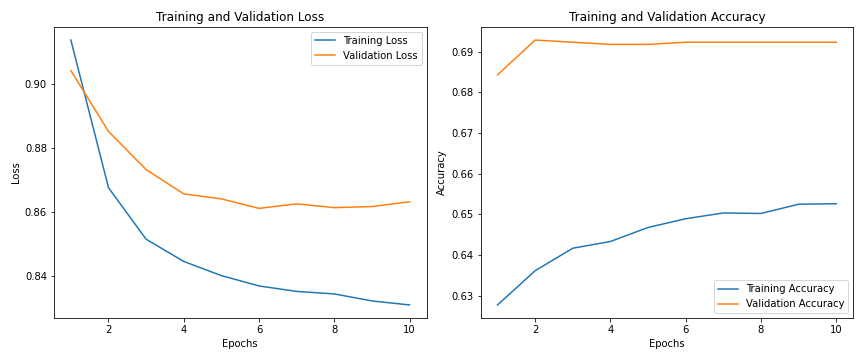}
    \caption{Training and validation loss and accuracy for audio model}
    \label{fig:audio_graph}
\end{figure}

The video model shows consistent improvement in training loss (Figure \ref{fig:video_graph}), but its validation loss slightly increases after an initial decrease, suggesting potential overfitting as the model learns. The text model exhibits a steady decrease in both training and validation losses with closely aligned curves, indicating effective learning and strong generalization (Figure \ref{fig:text_graph}). In contrast, the audio model demonstrates stable training loss reduction, but its validation loss plateaus early, highlighting challenges in generalizing to unseen data (Figure \ref{fig:audio_graph}). Overall, the text model performs most reliably, while the video and audio models face overfitting and generalization issues, respectively.

\subsubsection{Approach 0: Late Stage Feature Fusion Results}
This approach, which combined results using majority voting, yielded the lowest accuracy of \textbf{66.23\%}. By treating each modality independently, the relationships between the modalities, which are critical for sentiment analysis, are not captured. These results serve as a baseline for comparison between other implementations.   

\subsubsection{Approach 1: Early Stage Feature Fusion Results}
\begin{figure}[htp]
    \centering
    \includegraphics[width=8cm]{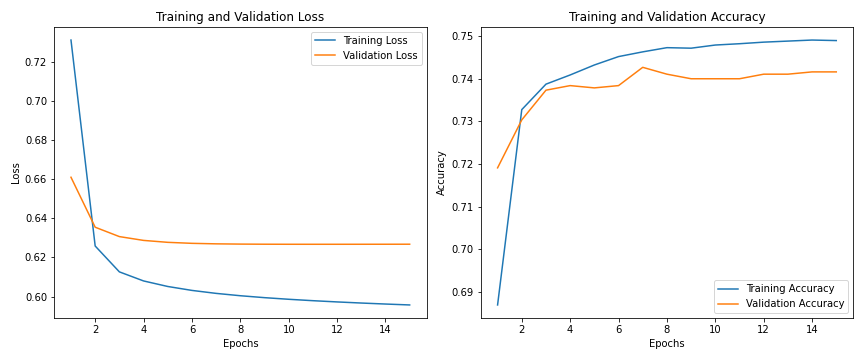}
    \caption{Training and validation loss and accuracy for early stage feature fusion model}
    \label{fig:approach1_graph}
\end{figure}

The pretrained multimodal fusion approach demonstrates a smooth decrease in training loss and an early stabilization of validation loss as shown in Figure~\ref{fig:approach1_graph}. Training and validation accuracies improve steadily and converge near 74\%, with minimal overfitting observed. The results highlight the effectiveness of feature fusion in combining modality-specific information, though gains plateau early, possibly due to limited model complexity.\\

By concatenating features from all modalities early and training a classifier on the combined feature representation, this approach outperformed Approach 0, with an accuracy of \textbf{71.87\%}. This indicates that combining the modalities early allows the model to learn a better representation of the features, thus improving the sentiment classification accuracy. 

\subsubsection{Approach 2: Multi-headed Attention Results}

\begin{figure}[htp]
    \centering
    \includegraphics[width=8cm]{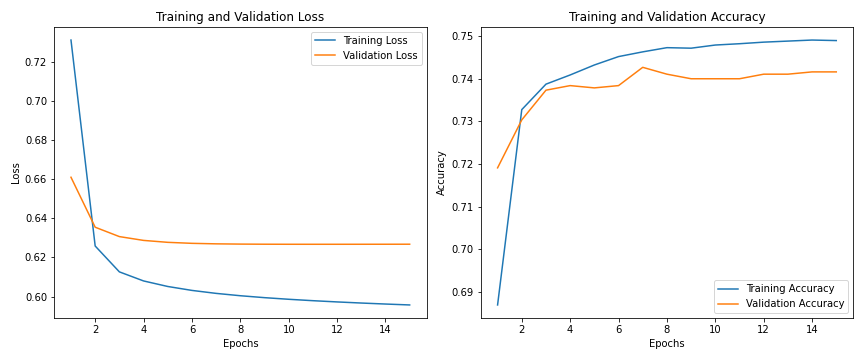}
    \caption{Training and validation loss and accuracy for multi-headed attention model}
    \label{fig:approach2_graph}
\end{figure}

The training and validation losses in this approach decrease steadily, with validation loss stabilizing towards the end, indicating effective learning without significant overfitting (Figure~\ref{fig:approach2_graph}). Both training and validation accuracies converge near 74\%, showing strong generalization. This suggests that the pretrained multimodal approach effectively leverages attention mechanisms and cross-modality interactions for robust performance.\\

This approach slightly improves on Approach 1, with an accuracy score of \textbf{72.39\%}, by incorporating attention mechanisms to weigh the importance of each modality's features. While the improvement in the accuracy is marginal over Approach 1, the addition of the attention layer is likely to provide more robustness where certain modalities are more critical to sentiment analysis. 
\section{Discussion and Future Work}
\subsection{Discussion}
The results from the different approaches demonstrated that integrating different modalities at an early stage improves the performance for sentiment analysis. The improvement of \textbf{5.64\%} from Approach 0 to Approach 1 highlighted that by combining features from text, audio and visual data, shared representation learning helped capture richer and more nuanced patterns.\\ 

Approach 2 builds on Approach 1 by introducing the attention mechanism to weigh the importance of each modality. While the accuracy improvement over Approach 1 was marginal, it can be suggested that the attention mechanism does not provide significant benefit under the current experimental setup, or that the current architecture and dataset might not be able to fully exploit the benefits of attention.\\ 

The results of the experiment underscore the limitations of using late-stage fusion approaches like Approach 0, which fail to leverage on the interactions between modalities and clearly shows the benefit of performing fusion early. 
\subsection{Future Work}
The following considerations can be taken for future work to further enhance the performance of multimodal sentiment analysis:
\begin{enumerate}
    \item Dataset Enrichment: Utilize datasets with greater modality-specific variation to better assess the impact of dynamic feature weighting from multi-headed attention.
    \item Fusion on Temporal Data: The dataset consisted of temporal information in the form of sequences. As early fusion has shown to be successful, by implementing fusion on the temporal sequences, the model could learn more nuanced and contextually aware feature representations.
\end{enumerate}

{\small
\bibliographystyle{ieee_fullname}
\bibliography{bib}

\begin{thebibliography}{1}\itemsep=-1pt

\bibitem{he2022DISFRN}
Jing He, Haonan Yanga, Changfan Zhang, Hongrun Chen, and Yifu Xua.
\newblock Dynamic invariant-specific representation fusion network for multimodal sentiment analysis.
\newblock {\em Computational Intelligence and Neuroscience}, 2022.

\bibitem{lai2023msasurvey}
Songning Lai, Xifeng Hu, Haoxuan Xu, Zhaoxia Ren, and Zhi Liu.
\newblock Multimodal sentiment analysis: A survey.
\newblock {\em Displays}, 80, 2023.

\bibitem{liang2023multizoo}
Paul~Pu Liang, Yiwei Lyu, Xiang Fan, Arav Agarwal, Yun Cheng, Louis-Philippe Morency, and Ruslan Salakhutdinov.
\newblock Multizoo \& multibench: A standardized toolkit for multimodal deep learning.
\newblock {\em Journal of Machine Learning Research}, 24:1--7, 2023.

\bibitem{rahman2020magbert}
Wasifur Rahman, Md~Kamrul Hasan, Sangwu Lee, Amir Zadeh, Chengfeng Mao, Louis-Philippe Morency, and Ehsan Hoque.
\newblock Integrating multimodal information in large pretrained transformers.
\newblock {\em Proc Conf Assoc Comput Linguist Meet}, Proceedings of the 58th Annual Meeting of the Association for Computational Linguistics:2359–--2369, 2020.

\bibitem{soleymani2017surveymsa}
Mohammad Soleymani, David Garcia, Brendan Jou, Björn Schuller, Shih-Fu Chang, and Maja Pantic.
\newblock A survey of multimodal sentiment analysis.
\newblock {\em Image and Vision Computing}, 65:3--14, 2017.

\bibitem{sun2021timf}
Jianguo Sun, Hanqi Yin, Ye Tian, Junpeng Wu, Linshan Shen, and Lei Chen.
\newblock Two-level multimodal fusion for sentiment analysis in public security.
\newblock {\em Security and Communication Networks}, 2021.

\bibitem{wang2022tedt}
Fan Wang, Shengwei Tian, Long Yu, Jing Liu, Junwen Wang, Kun Li, and Yongtao Wang.
\newblock Tedt: Transformer-based encoding–decoding translation network for multimodal sentiment analysis.
\newblock {\em Cognitive Computation}, 1, 2023.

\end{thebibliography}
}

\end{document}